\documentclass[runningheads]{llncs}
\usepackage[T1]{fontenc}
\usepackage{graphicx}
\usepackage{amsmath}
\usepackage{amssymb}
\usepackage{booktabs}
\usepackage{amsmath}
\usepackage{subcaption}
\usepackage{algorithm}
\usepackage{wrapfig}
\usepackage{pifont}
\usepackage{sidecap} 
\usepackage[breaklinks,colorlinks]{hyperref}

\begin{document}
\title{Linked Adapters: Linking Past and Future to Present for Effective Continual Learning}
\titlerunning{Linked Adapters for Continual Learning}
\author{}
\institute{}

%
\author{%
 Dupati Srikar Chandra$^{1}$, P. K. Srijith$^{1}$, Dana Rezazadegan$^{2}$, Chris McCarthy$^{2}$\\
\email{$^{1}$Indian Institute of Technology Hyderabad, India} \\
\email{$^{2}$Swinburne University of Technology, Australia}\\
{\tt\small \{ai20resch11004\}@iith.ac.in, srijith@cse.iith.ac.in,} \\ {\tt\small\{drezazadegan, cdmccarthy\}@swin.edu.au
}}

\authorrunning{D. S. Chandra et al.}

\maketitle              

\begin{abstract}
Continual learning allows the system to learn and adapt to new tasks while retaining the knowledge acquired from previous tasks. However, deep learning models suffer from catastrophic forgetting of knowledge learned from earlier tasks while learning a new task. Moreover, retraining large models like transformers from scratch for every new task is costly. An effective approach to address continual learning is to use a large pre-trained model with task-specific adapters to adapt to the new tasks. Though this approach can mitigate catastrophic forgetting, they fail to transfer knowledge across tasks as each task is learning adapters separately. To address this, we propose a novel approach \textit{Linked Adapters} that allows knowledge transfer through a weighted attention mechanism to other task-specific adapters. Linked adapters use a multi-layer perceptron (MLP) to model the attention weights, which overcomes the challenge of backward knowledge transfer in continual learning in addition to modeling the forward knowledge transfer. During inference, our proposed approach effectively leverages knowledge transfer through MLP-based attention weights across all the lateral task adapters. Through numerous experiments conducted on diverse image classification datasets, we effectively demonstrated the improvement in performance on the continual learning tasks using \textit{Linked Adapters}.
\end{abstract}

\section{Introduction}

In various applications, computational systems encounter the challenge of learning from a continuous stream of data and dynamically adjusting to their environment based on knowledge acquired from previous experiences. For example, an autonomous agent navigating the real world must be able to continually learn from a stream of data, ensuring that they retain the knowledge gained from diverse and ever-shifting distributions without the threat of forgetting knowledge. Despite the remarkable performance of deep learning models in computer vision \cite{dosovitskiy2020image,krizhevsky2017imagenet}, they still struggle with the challenge of learning multiple tasks sequentially without forgetting previously acquired knowledge. This phenomenon of forgetting knowledge acquired from previous tasks while attempting to learn new tasks is known as catastrophic forgetting \cite{mccloskey1989catastrophic}. Achieving the capability to continuously learn from a continuous stream of tasks without compromising the retention of knowledge from prior tasks, all the while accommodating new tasks referred to as continual learning (CL)~\cite{parisi2018continual,mermillod2013stability,grossberg2007consciousness}.

Vision transformers (ViTs) \cite{dosovitskiy2020image,carion2020end} have proven effective for computer vision tasks like classification, object detection, and segmentation. However, Transformer-based models generally require larger datasets and longer training times to achieve their full potential, mainly due to their extensive parameter count. To address this, various approaches \cite{dai2015semi,howard2018universal,radford2018improving} have leveraged pre-trained large models (PLMs) such as ViT \cite{dosovitskiy2020image} and DeiT \cite{deit}, fine-tuning them on new tasks. In CL scenarios, where tasks are presented one after another, fine-tuning can overwrite pre-trained knowledge, posing a challenge in adapting PLMs to new tasks while preventing forgetting and controlling parameter growth.



An effective approach to bringing continual learning capability in PLMs is through adapters~\cite{houlsby2019parameter,rebuffi2018efficient}. Adapters are small trainable parameters that are added to a pre-trained network to enable it to adapt to new tasks. To perform CL, a pre-trained network is typically frozen, and only the task-specific adapter parameters are learned for each new task which allows it to retain information from prior tasks. Approaches such as \cite{houlsby2019parameter} have shown the efficacy of adapters in fine-tuning downstream tasks, showcasing their effectiveness in mitigating forgetting. The major limitation of the approach is not being able to transfer knowledge gained from one task to another because the model is frozen, and the task-specific parameters are learned independently.

To address the aforementioned limitations, we propose a novel approach, \textit{Linked Adapters} that enables knowledge transfer across tasks in both forward and backward directions. In \textit{Linked Adapters}, we maintain a  weighted attention-based lateral connections across task-specific adapter representations which enable knowledge transfer across tasks. Specifically, these attention weights help to determine the degree of information flow from other tasks to the particular task through the adapter representations. A major challenge here is to learn the attention weights from other task-specific adapter representations in a CL setup, where we see tasks in a sequence. Consequently, it is difficult to learn the attention weights from subsequent tasks as the adapter representations from subsequent tasks are not available while learning a task. To address this challenge, we learn to predict attention weights using a \textit{Multi-layer Perceptron (MLP)}. The \textit{MLP} is trained to predict the attention weights from the embeddings of the seen tasks. During the testing phase, the \textit{MLP} can be used to predict the attention weights from all other tasks given the task embeddings.
 The proposed approach allows both forward and backward knowledge transfer across task-specific adapters and helps the initial tasks to benefit from the knowledge learned from the subsequent tasks as well. Our experimental results demonstrate the effectiveness of \textit{Linked Adapters} in transferring knowledge across tasks on various image classification datasets.

Our main contributions are summarised as follows:
\vspace{-0.1cm}
\begin{itemize} 
\item
We propose a novel approach \textit{Linked Adapters} that transfers the knowledge using a weighted attention mechanism to other tasks.
\item We develop a \textit{MLP} based attention weight generation in \textit{Linked Adapters}  that enables backward transfer of knowledge to initial tasks without any extra training time.
\item We empirically show that \textit{Linked Adapters} can transfer knowledge across tasks on several image classification datasets in both forward and backward directions.
\end{itemize}

\section{Related Work}

Catastrophic forgetting is a critical challenge in continual learning, where neural networks must adapt to new tasks without losing previously learned knowledge. Several strategies have been proposed to mitigate this issue, broadly categorized into Regularization-based approaches \cite{aljundi2018memory,kirkpatrick2017overcoming,zenke2017continual}, Replay based approaches~\cite{lopez2017gradient,shin2017continual,rebuffi2017icarl} and Dynamic architecture based approaches \cite{yoon2017lifelong,rusu2016progressive,dytox}. Regularization methods, such as EWC \cite{kirkpatrick2017overcoming} and SI \cite{zenke2017continual}, constrain important parameters of previous tasks while allowing new tasks to adapt. Replay methods like DGR \cite{shin2017continual} and iCaRL \cite{rebuffi2017icarl} store or synthesize samples from prior tasks for replay. Dynamic architecture methods, such as PNNs \cite{rusu2016progressive} and DEN \cite{yoon2017lifelong} , expand network architecture to accommodate new tasks.



Pre-trained models have shown a significant impact on transfer learning across various tasks with faster convergence and reduced computational costs \cite{dai2015semi,howard2018universal,radford2018improving}. Usually, pre-trained models like DeiT \cite{deit} and ViT \cite{dosovitskiy2020image}, are fine-tuned for specific downstream tasks to leverage their existing knowledge. However, in continual learning, updating the entire network for each new task can cause catastrophic forgetting. Various approaches~\cite{houlsby2019parameter,rebuffi2018efficient} have been proposed to update a few task-specific parameters for each downstream task instead of finetuning on all the parameters. Recently, transformer-specific adapters~\cite{houlsby2019parameter} are proposed to learn downstream tasks by updating only a few tasks-specific parameters. ADA~\cite{ermis2022continual} selects and distils adapters from a pool for new tasks. However, as the number of tasks increases, shared adapters may suffer from forgetting.

In recent advancements, prompt-based approaches \cite{l2p,dualprompt,coda,jung2023generating,ranpac}, have shown promising results in adaptively leveraging pre-trained knowledge for downstream continual learning. L2p~\cite{l2p} employs a pool of prompts with the corresponding keys and, during inference, the most relevant keys and the corresponding prompts are selected. Dual-Prompt~\cite{dualprompt} maintains task-sharing and task-specific prompts whereas CODA-Prompt~\cite{coda} learns a set of prompt components through weighted summation. DAP-CL \cite{jung2023generating} introduces Domain-Adaptive Prompts that generate instance-level prompts at inference time.

\vspace{-0.25cm}
\section{Background}
\subsection{Problem Formulation}

Continual learning assumes tasks to arrive in a sequential manner, and there exists a limited memory so we cannot store the previous task data or model parameters. Our goal is to effectively learn a set of \(m\) image classification tasks, denoted as \( \mathcal{T}=\{\mathcal{T}^{1},\mathcal{T}^{2},\ldots,\mathcal{T}^{m}\} \), as they arrive in sequence. Each task \(t\in\mathcal{T}\) is associated with data  \((X^{t}, Y^{t})\), where \(X^{t}=\{x_{i}^{t}\}_{i=1}^{N^t}\) represents the input data points with $x_{i}^{t} \in \mathcal{X}$, and \(Y^{t}=\{y_{i}^{t}\}_{i=1}^{N^t}\) represents the corresponding output data points with $y_{i}^{t} \in \mathcal{Y}^t$. \(N^t\) represents the number of samples in the task $t$ and let $\mathcal{Y} = \{\mathcal{Y}^1, \ldots, \mathcal{Y}^m \}$ represent the set of classes in all tasks. Our objective is to learn a model $f{(\cdot, {\Theta}}):\mathcal{X}\to\mathcal{Y}$ with parameter $\Theta$ that can provide a good generalization performance across all tasks.

\subsection{Adapters}

Training large models like transformers from scratch on new tasks is time-consuming and requires larger datasets to train. An efficient way is to use pre-trained models, which can be fine-tuned on downstream tasks to transfer knowledge and often leads to faster convergence. In the continual learning scenario, task arrives sequentially, and finetuning parameters on each new task result in losing information on pre-trained data and earlier tasks. Adapters \cite{houlsby2019parameter} address this by adding task-specific parameters, $\Phi^t$, without modifying the pre-trained model weights ($\Theta$). Adapter $(\mathcal{A}^{t})$ typically uses a bottleneck architecture with MLP layers that downsample $(D^{t})$ and upsample $(U^{t})$ representations. For a downstream task $t$, adapters are inserted at each layer of the transformer, where the hidden representation is transformed as $\tilde{h}^{t}_{k} = U_{k}^t(D_{k}^{t}(\bar{h}^{t}_{k}))$. Here, $\bar{h}^{t}_{k}$ and $\tilde{h}^{t}_{k}$ are hidden representation input and output at layer $k$. The model learns only the adapter parameters $\Phi^t$ for each task, along with task-specific classifier head ($\Psi^t$), keeping $\Theta$ frozen. As pre-trained model weights are frozen, learning separate adapters independently won't transfer knowledge across tasks. In a continual learning scenario, we need a model that transfers knowledge across tasks.
\vspace{-0.75cm}
\section{Linked Adapters}
In a continual learning scenario, where tasks are learned sequentially, knowledge gained from one task can be leveraged to facilitate learning in subsequent tasks if tasks share common underlying representations. However, a significant limitation arises when employing separate and independent adapters \cite{houlsby2019parameter}, as this approach involves freezing the pre-trained model and restricting learning to task-specific parameters in isolation. To address this limitation, we aim to propose \textit{Linked Adapters}, that can transfer knowledge across tasks by maintaining systematic lateral connections to the adapters from all other tasks. The challenge arises in determining the degree of information flow required, as information should ideally flow more prominently between similar tasks and less among dissimilar tasks. To address this, we utilize attention weights to control the flow of information tailored to each task. In a CL scenario, where tasks arrive sequentially, learning attention weights becomes even more challenging since all tasks and their corresponding adapters are not available at once. To overcome this, we employ a Multi-Layer Perceptron (MLP) trained to generate attention weights of lateral connections from previous task adapter representations. 
During testing, the \textit{MLP} proves invaluable for leveraging the knowledge gained from subsequent tasks by predicting the attention weights from the subsequent tasks. This benefits the initial tasks that receive limited information otherwise from the tasks prior to them without requiring any additional training. In the following sections, we will discuss in detail the working of \textit{Linked Adapters}.

\subsection{Linked Adapter Training }
\label{linked_adapter_training}
\begin{figure*}[t]
  \centering
    \includegraphics[width=1\linewidth, height=6cm]{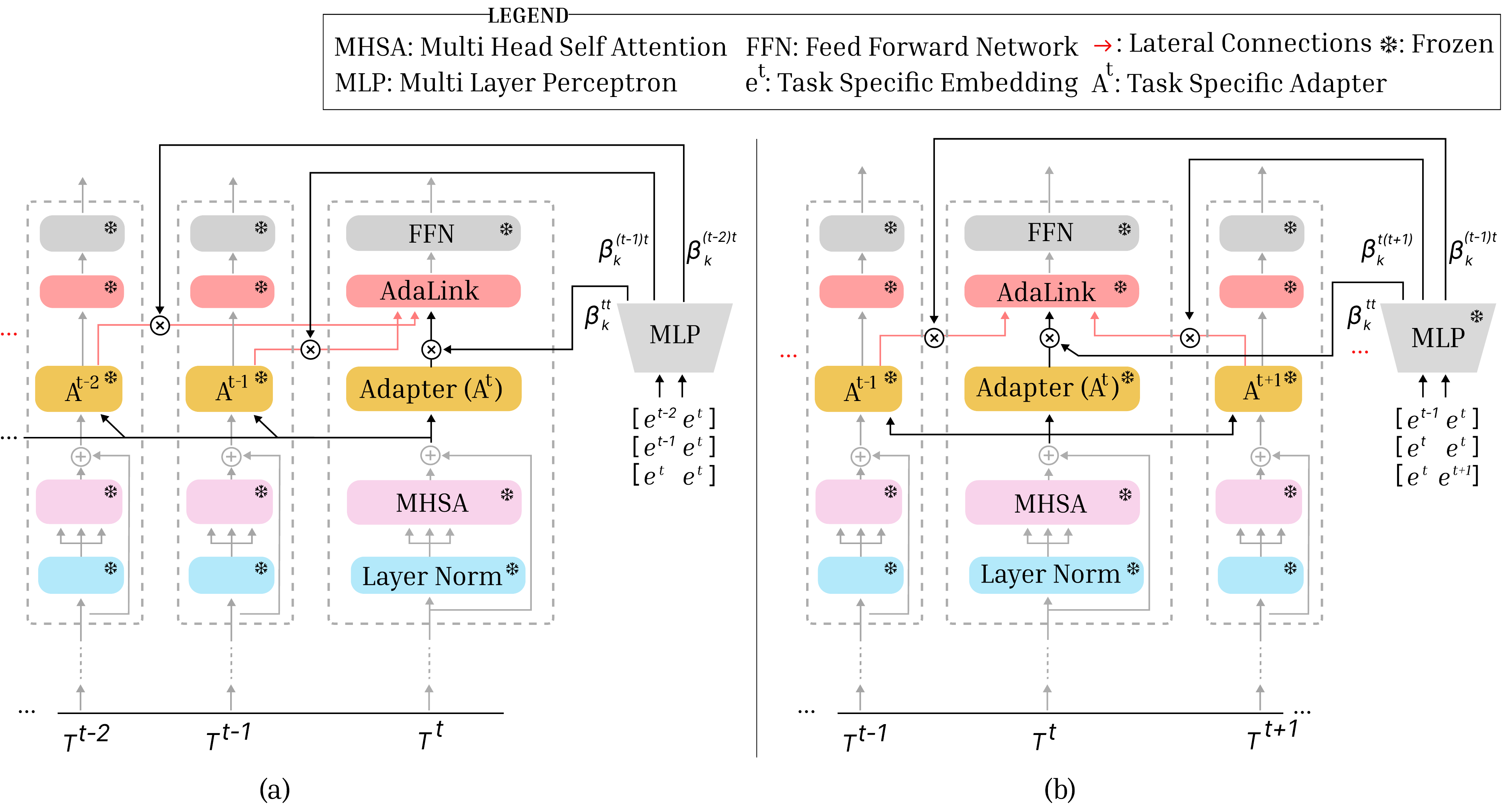}
     \caption{Figure \ref{fig:architecture}.a. demonstrates \textit{Linked Adapters} during training where lateral connections from previous task adapter representations are connected to current task $t$ to enable knowledge transfer and their corresponding attention weights are being generated by \textit{MLP} . Figure \ref{fig:architecture}.b. demonstrates \textit{Linked Adapters} during testing where the lateral connections from the adapter representations from previous and subsequent tasks are added to current task $t$ enable knowledge transfer from both directions where \textit{MLP} generates attention weights of subsequent tasks without any extra training.}
     \label{fig:architecture}
     \vspace{-0.5cm}
\end{figure*}

During training, we see the tasks in a sequence, \textit{Linked Adapters} maintains lateral connections, which establish links between the representations of previous task adapters to the current learning adapter as in Figure~\ref{fig:architecture}.a. A \textit{MLP} $f_{h}(.,\Theta_{h})$ is learned along with task-specific adapter parameters to generate forward attention weights between every previous task to current task. These attention weights ensure only useful information is transferred to the current task. To generate forward attention weights $\beta^{pt}$ between previous task $(p)$ adapter representation to current task $t$, \textit{MLP} takes task-specific embeddings of previous task $\mathbf{e}^p$ and task embedding of current task $\mathbf{e}^t$ as input. Task embeddings are small, trainable parameters that help in capturing task-specific information in Linked Adapters. The forward attention weight generation can be defined as  
\begin{equation}\label{eq:1}
 \beta^{pt}= \{\beta^{pt}_k\}_{k=1}^{L}=
f_{h}(\mathbf{e}^p,\mathbf{e}^t;\Theta_{h})  
\end{equation}
where $k$ denotes layer number of transformer with $L$ layers, $\mathbf{e}^p$ is frozen and only current task embedding $\mathbf{e}^t$ is learned along with \textit{MLP} and stored at the end of each task. These stored embeddings are used as input to \textit{MLP} to generate attention weights while learning new tasks. Here, $\Theta_h$ are the parameters of the \textit{MLP} that are optimized during training to model the attention weights. 

\textit{Linked Adapters} allows transfer of knowledge from previous tasks to the current task, by maintaining lateral connections at every layer of the transformer network, and the intention of having weighted attention to the previous task-specific adapter representations is to ensure that only useful information is transferred to the current task. For current task \textit{t}, weighted lateral connections from previous task-specific adapter representations to the current task at $k^{th}$ layer of transformer using \textit{Linked Adapters} can be defined as:

\begin{equation}\label{eq:3}
{\tilde{h}}^{t}_{k}= \sum_{p=1}^{t-1}  \beta^{pt}_{k} U^{p}_{k}(D^{p}_{k}( \bar{h}^{t}_{k}))+ \beta^{tt}_{k} U^{t}_{k}(D^{t}_{k}( \bar{h}^{t}_{k}).
\end{equation}

Here, $ \bar{h}^{t}_{k}$ is hidden state input to the adapter, $\beta^{pt}_{k}$ are the attention weights computed between previous task $p$ and current task $t$ which determines the amount of information to keep from the previous task $p$. $ {\tilde{h}}^{t}_{k}$ is a new representation obtained by considering the weighted attention computed over adapter representations from the previous tasks. 

We provide below the sequence of operations happening inside a transformer block in our model. Let $h^{t}_{k-1}$ be the hidden representation output of $(k-1)^{th}$ transformer layer and is provided as input to $k^{th}$ layer. The following operations are performed on the $k^{th}$ layer of the transformer block.
\begin{align*}
    &h_k^{'t} = {h}^{t}_{k-1} + \text{MHSA}(\text{Norm}({h}^{t}_{k-1})), \quad 
    \bar{h}_k^{t}= \text{Norm}(h_k^{'t}))\\
    &\tilde{h}_k^{t} = \sum_{p=1}^{t-1} \beta_k^{pt} U_k^p(D_k^p(\bar{h}_k^{t})) + \beta_k^{tt} U_k^t(D_k^t(\bar{h}_k^{t})), \hspace{0.5em} 
    \hat{h}_k^{t} = \bar{h}_k^{t} + \tilde{h}_k^{t}, \hspace{0.5em} h^{t}_{k} = \bar{h}_k^{t} + \text{FFN}(\hat{h}_k^{t})
\end{align*}
where Norm, MHSA, and FFN are the layer normalization, Multi-Headed Self Attention, and Feed Forward layers of $k^{th}$ layer of the transformer respectively.

The proposed model includes some pre-trained model parameters which are frozen ($\Theta^{\dagger}$), while task-specific parameters such as adapter parameters ($\Phi^{t}$), task-embeddings ($\mathbf{e}^t$) and classification head parameters ($\Psi^t$) are learned from the corresponding task data. The MLP parameters ($\Theta_{h}$) are updated on the data from all the tasks as we see them in the sequence. The learning of parameters in the \textit{Linked Adapters} while training on data from task $t\in\mathcal{T}$ is given as :
 \begin{equation}
 \label{eq:taskloss}
    \operatorname*{arg max}_{\Phi^{t},\mathbf{e}^t,\Theta_h,\Psi^t} \sum_{i=1}^{N^t} \log \mathcal{P}(y^t_i | x^t_i, \Theta^{\dagger},\Theta_{h},\mathbf{e}^t,\Phi^t, \Psi^t)
\end{equation}


While attempting to learn on a new task, \textit{MLP} may overfit the current task, which could cause it to lose the knowledge gained from previous tasks. To address this, we aim to regularize the \textit{MLP} parameters so that the MLP should be able to generate attention weights with respect to the previous tasks as well. The regularization over the \textit{MLP} parameters retains the values of the important \textit{MLP} parameters associated with the previous task learning stages similar to an Online-EWC~\cite{onlineewc}.  While learning the current task, the objective is to optimize the total loss which includes the task-specific loss (negative of the objective in eq.~\ref{eq:taskloss}) and the regularization term, and is defined as
\begin{equation}\label{eq:4}
\small
\operatorname*{arg min}_{\Theta_{h},\Phi^t,\Psi^t,\mathbf{e}^t} \mathcal{L}_{total}  =  \mathcal{L}_{task}(X^{t},Y^{t},\Theta_{h},\Theta^\dagger,\mathbf{e}^{t},\Phi^t,{\Psi^t})  +{  {\lambda} \sum_{j}
 FI_{j}^{t-1}{(\Theta_{h,j}^{*} -\Theta_{h,j})^2 }}.
\end{equation}

Here, the first term
\(\mathcal{L}_{task}\) is task-specific loss and the second term helps to regularise the \textit{MLP} weights to avoid forgetting, and $\lambda$ is the regularization constant. $\Theta_{h}^ {*}$ are the \textit{MLP} weights after learning previous task and $\Theta_{h}$ are the \textit{MLP} weights that we are learning on the current task $t$. $FI$ indicates fisher information matrix that computes the importance of weights $\Theta_{h}^ {*}$ in \textit{MLP}, and the index $j$ iterates over all the \textit{MLP} parameters.


\subsection{Inference with Linked Adapters }
\label{linked_adapter_inference}
During training time, we considered only forward connections as the adapter parameters from the subsequent tasks are yet to be learned due to the CL setup. During inference time, we have access to all the task-specific adapters (parameters and not data). Hence, we can transfer knowledge acquired in the adapters from all other tasks to a particular task through our weighted connections. This can help all the tasks  to improve their performance through knowledge transfer, and especially initial tasks that may not have received sufficient knowledge through lateral connections from subsequent tasks. 

During the testing phase for the current task $t$, the model effectively employs lateral connections to the adapter representations from both prior and subsequent tasks to $t$, enabling knowledge transfer in both forward and backward directions, as shown in  Figure~\ref{fig:architecture}.b. However, a challenge arises when generating attention weights for these lateral connections from subsequent tasks, as the attention weights and the degree of information flow from these tasks cannot be learned during the training phase. We address this challenge using a \textit{MLP} learned during the training phase. \textit{MLP} facilitates the generation of attention weights for lateral connections from subsequent tasks to the current task without requiring additional training. To achieve this, the \textit{MLP} takes task-specific embeddings of subsequent tasks as input, which are learned during the training phase, along with the current task embeddings and generates attention weights.

During testing the task at hand $t$, after training on all $\mathcal{T}$ tasks, \textit{MLP} can generate forward attention weights $\beta^{pt}$ from the previous task \textit{p} to task \textit{t}  and  also backward attention weights  $\beta^{ts}$ from subsequent task \textit{s}  to current task \textit{t}  and are  defined as 

\begin{equation}\label{eq:5}
\begin{aligned}
 & \beta^{pt}= \{\beta^{pt}_k\}_{k=1}^{L}=
f_h(\mathbf{e}^p,\mathbf{e}^t;\Theta_{h}), \quad
\beta^{ts} = \{\beta^{ts}_k\}_{k=1}^{L} = f_h(\mathbf{e}^t, \mathbf{e}^s; \Theta_{h})
\end{aligned}
\end{equation}
 For task $t$, weighted lateral connections using Linked Adapters from both previous and subsequent tasks at $k^{th}$ layer using \textit{MLP} generated attention weights can be defined as:
\begin{equation}\label{eq:7}
\footnotesize
{\tilde{h}}^{t}_k=\sum_{p=1}^{t-1}\beta^{pt}_{k} U^{p}_{k}(D^{p}_{k}( \bar{h}^{t}_{k}))
        +\beta^{tt}_{k} U^{t}_{k}(D^{t}_{k}( \bar{h}^{t}_{k})+\sum_{s=t+1}^{m}\beta^{ts}_{k} U^{s}_{k}(D^{s}_{k}( \bar{h}^{t}_{k}))
\end{equation}
Using these backward connections, subsequent tasks can transfer knowledge to initial tasks in the backward direction.

Using \textit{ Linked Adapters}, the pre-trained model can be fine-tuned for multiple tasks without requiring significant additional training. This approach is especially useful when dealing with tasks that share some underlying structure or knowledge, as it allows the model to leverage this shared information to improve performance on multiple tasks. Specifically,  our \textit{Linked Adapters} is compatible with all transformer-based architectures.

\section{ Experiments}
We performed extensive experiments on various image classification benchmarks to show
the effectiveness of our approach. We consider two variants of our proposed approach to conduct experiments. 
\begin{itemize} 
\item \textit{AdaLink-Bidirectional} : the proposed  \textit{Linked Adapters} which allow knowledge transfer from previous and subsequent tasks. It considers weighted lateral connections from previous and subsequent tasks at the test time. 
\item \textit{AdaLink-Forward} : the variant of the  \textit{Linked Adapters} which allow knowledge transfer only  from previous tasks. This model considers weighted lateral connections only from previous tasks to the current tasks at the test time. 
\end{itemize}

\textbf{Datasets}: We present our results on popular image classification datasets, \textit{CIFAR100}~\cite{cifar100}, \textit{CUB200} ~\cite{cub200} and Imagenet-R \cite{imagenet-r} datasets. The proposed approach was evaluated on the \textit{Split-CIFAR100} dataset, splitting into 5, 10, and 20 tasks to demonstrate the approach's adaptability to various task sequences. Further evaluations were conducted on the \textit{CUB200} dataset, containing images of 200 bird subcategories, with splits of 5 and 10 tasks, as well as on \textit{Imagenet-R} with 200 classes divided into 10 tasks.



\textbf{Experimental Setup}: We conducted experiments in task incremental setup where the task identity is provided both during training and test time. For our experiments, we used the pre-trained Vision Transformer ($ViT\_B\_16$) as the base network.

\textbf{Baselines}: The approach we proposed is being evaluated against several established baseline methodologies in the field of continual learning, including \textit{Finetuning}, where model parameters adapt to new tasks without constraints; \textit{EWC} \cite{kirkpatrick2017overcoming}, which imposes constraints on model parameters important to old tasks while allowing others to adapt to new tasks; \textit{Experience Replay (ER)} \cite{experience}, involving storing and replaying samples from previous tasks while learning new ones; and \textit{Adapters} \cite{houlsby2019parameter}, which learns separate and independent adapter modules for each task without knowledge transfer or forgetting. Additionally, results reported for L2P \cite{l2p}, Dual-Prompt \cite{dualprompt}, CODA-Prompt \cite{coda} and DAP-CL\cite{jung2023generating} in Table \ref{table:1} are in a task-incremental setup where task identity is provided during inference to select a task-specific classifier head.

\textbf{Evaluation Metrics}: We use the standard evaluation metric of computing the averaged accuracy across all tasks after sequential training on the tasks.  To validate the effectiveness of \textit{AdaLink} in enabling knowledge transfer over a sequence of tasks, we compute a \textit{Knowledge Transfer(KT)} metric. For \textit{AdaLink}, the average knowledge transfer across all tasks can be computed as $KT=\frac{1}{m}\sum_{i=1}^{m} {acc}_{link,i} - {acc}_{a,i} $ where ${acc}_{link,i}$ is the accuracy of $i^{th}$  task using \textit{AdaLink}, ${acc}_{a,i}$ is the accuracy of the task $i^{th}$ trained using a separate and independent adapter approach.

\begin{table}
\caption{Comparison of average test accuracy ($\%$) across various splits of \textit{Cifar-100, CUB-200 and Imagenet-R} datasets (computed over 5 random seed initializations).
}

\resizebox{\textwidth}{!}{%
\begin{tabular}{|p{3.3cm}|c|c|c|c|c|c|}
\hline
\multicolumn{1}{|c|}{} &
  \multicolumn{3}{c|}{Split-Cifar-100} &
  \multicolumn{2}{c|}{Split-CUB-200}& \multicolumn{1}{c|}{Split-Imagenet-R}  \\ \hline
 
&
  5 tasks &
   10 tasks &
  20 tasks &
  5 tasks &
  10 tasks &  10 tasks  \\ \hline
\textit{Finetuning}   & 92.64  \scriptsize$\pm$ 0.57 &94.44  \scriptsize$\pm$ 0.49 &  96.25  \scriptsize$\pm$ 0.59   & 77.93  \scriptsize$\pm$  0.41 &  77.64  \scriptsize$\pm$ 0.68 & 76.98 \scriptsize$\pm$ 0.43  \\

\textit{EWC}          &  92.89  \scriptsize$\pm$ 0.40 &  95.03  \scriptsize$\pm$ 0.45 &  96.42  \scriptsize$\pm$ 0.96   & 78.02  \scriptsize$\pm$ 0.95 &  78.07  \scriptsize$\pm$ 0.58& 78.28 \scriptsize$\pm$ 0.29\\

\textit{ER}         & 93.27  \scriptsize$\pm$ 0.42 & 95.69  \scriptsize$\pm$ 0.50 & 97.19  \scriptsize$\pm$ 0.31 & 78.94  \scriptsize$\pm$ 0.94 & 78.31  \scriptsize$\pm$ 0.83&   78.45 \scriptsize$\pm$ 0.38\\

\textit{Adapters}    & 94.41  \scriptsize$\pm$ 0.10 & 96.57  \scriptsize$\pm$ 0.24 & 97.99  \scriptsize$\pm$ 0.14   & 80.81  \scriptsize$\pm$ 1.22 & 80.51  \scriptsize$\pm$ 1.28&  85.31 \scriptsize$\pm$ 0.31  \\

\textit{L2P}&90.47 \scriptsize$\pm$ 0.34&93.36 \scriptsize$\pm$ 0.68&96.25 \scriptsize$\pm$ 0.13&74.66 \scriptsize$\pm$ 0.81& 77.61 \scriptsize$\pm$ 0.61& 72.95 \scriptsize$\pm$ 0.67\\

\textit{Dual-Prompt}&93.41 \scriptsize$\pm$ 0.25&96.33 \scriptsize$\pm$ 0.35&98.03 \scriptsize$\pm$ 0.11&75.56 \scriptsize $\pm$ 1.50 & 78.00 \scriptsize$\pm$ 2.33& 83.50 \scriptsize$\pm$ 0.47\\

\textit{CODA-Prompt}  &93.92 \scriptsize$\pm$ 0.10&96.68 \scriptsize$\pm$ 0.14&98.10 \scriptsize $\pm$ 0.19&76.04 \scriptsize$\pm$ 0.48& 79.23 \scriptsize$\pm$1.30&85.36 \scriptsize$\pm$ 0.33\\

\textit{DAP-CL} & 94.11 \scriptsize$\pm$ 0.16 & 96.53 \scriptsize$\pm$ 0.17&97.92 \scriptsize$\pm$ 0.16& 74.59 \scriptsize$\pm$ 1.03 &  78.90 \scriptsize$\pm$ 0.71 & 79.65 \scriptsize$\pm$ 0.93\\
\hline
\textit{AdaLink-Forward}  & 95.66  \scriptsize$\pm$ 0.18 &97.39  \scriptsize$\pm$ 0.16 & 98.50  \scriptsize$\pm$ 0.14   &  82.46  \scriptsize$\pm$ 1.21 & 80.94  \scriptsize$\pm$ 1.12 & 85.83 \scriptsize$\pm$ 0.36\\

 \textit{AdaLink-Bidirectional}  & \textbf{95.96  \scriptsize$\pm$ 0.31} & \textbf{97.73  \scriptsize$\pm$ 0.21}  & \textbf{98.57  \scriptsize$\pm$ 0.18}    & \textbf{82.76  \scriptsize$\pm$ 1.03} & \textbf{81.59  \scriptsize$\pm$ 0.83}& \textbf{86.16 \scriptsize$\pm$ 0.41}\\ \hline
\end{tabular}%
}
\label{table:1}
\end{table}

\begin{table}
\vspace{-0.5cm}
\caption{ Average Knowledge Transfer (\%) across tasks in AdaLink-Forward and AdaLink-Bidirectional in comparison to standalone adapters, where it has no knowledge transfer. 
}
\centering
\begin{tabular}{|p{3.3cm}|c|c|c|c|c|c|}
\hline
\multicolumn{1}{|c|}{} &

  \multicolumn{3}{c|}{Split-Cifar-100} &
  \multicolumn{2}{c|}{Split-CUB-200} & \multicolumn{1}{c|}{Split-Imagenet-R} \\ \hline
 &

  5 tasks &
  10 tasks &
  20 tasks &
  5 tasks &
  10 tasks &
  10 tasks 
  \\ \hline
\textit{AdaLink-Forward}    &1.25&0.82&0.51& 1.65&0.43& 0.52\\
\hline
 \textit{AdaLink-Bidirectional} &1.55&1.16&0.58&1.95&1.08& 0.85\\ \hline
\end{tabular}%
\label{table:2}
\end{table}

\vspace{-1cm}
\section{ Results}

We compare the performance of the proposed approach with the baselines on the image classification datasets, \textit{CIFAR100}, \textit{CUB200} and \textit{Imagenet-R} datasets in Table~\ref{table:1}. 
In experiments on the Split-CIFAR100 dataset, the \textit{AdaLink-Forward} method consistently outperformed baseline models across various splits. When tested with 5 splits, \textit{AdaLink-Forward} demonstrated a significant 1.25\% improvement in knowledge transfer compared to standalone adapters, leveraging lateral connections with \textit{MLP}-based weighted attention, as illustrated in Table~\ref{table:2}. Meanwhile, \textit{AdaLink-Bidirectional}, which incorporates both forward and backward task connections, surpassed \textit{AdaLink-Forward} by achieving a 1.55\% improvement in knowledge transfer, utilizing \textit{MLP}-based attention to enhance information flow from subsequent tasks. Performance gains were similarly observed with 10 and 20 splits, where \textit{AdaLink-Forward} improved knowledge transfer by 0.82\%  and 0.51\% respectively, while \textit{AdaLink-Bidirectional} showed improvements of 1.16\% and 0.58\%, as in Table~\ref{table:2}.

    



On the Split-CUB200 dataset, \textit{AdaLink-Forward} demonstrated its effectiveness across both 5 and 10 splits by outperforming all baseline models, showing substantial improvements over standalone adapters of 1.65\% and 0.43\%, respectively, as seen in Table~\ref{table:1}. Meanwhile, \textit{AdaLink-Bidirectional}, utilizing both forward and backward connections, achieved notable knowledge transfer gains of 1.95\% and 1.08\% over adapters in the respective splits, emphasizing its robust ability to transfer knowledge across tasks.

On the Imagenet-R dataset, \textit{AdaLink-Forward} and \textit{AdaLink-Bidirectional} approaches outperformed all baseline models on 10 splits, reported in Table \ref{table:1}. Compared to Standalone adapters, \textit{AdaLink-Forward} and \textit{AdaLink-Bidirectional} have shown improvement of 0.52\% and 0.85\% as reported in Table~\ref{table:2}.

\begin{figure*}
\vspace{-0.5cm}
    \centering
    
    \begin{subfigure}{0.49\textwidth}
        \includegraphics[width=\linewidth, height=4.5cm]{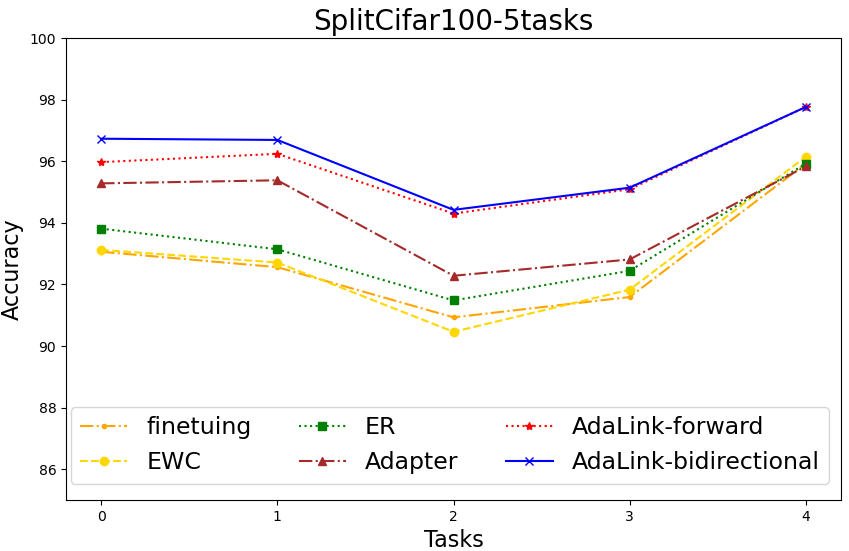}
        \caption{}
        \label{fig:sub1}
    \end{subfigure}
    \hfill
    \begin{subfigure}{0.49\textwidth}
        \includegraphics[width=\linewidth, height=4.5cm]{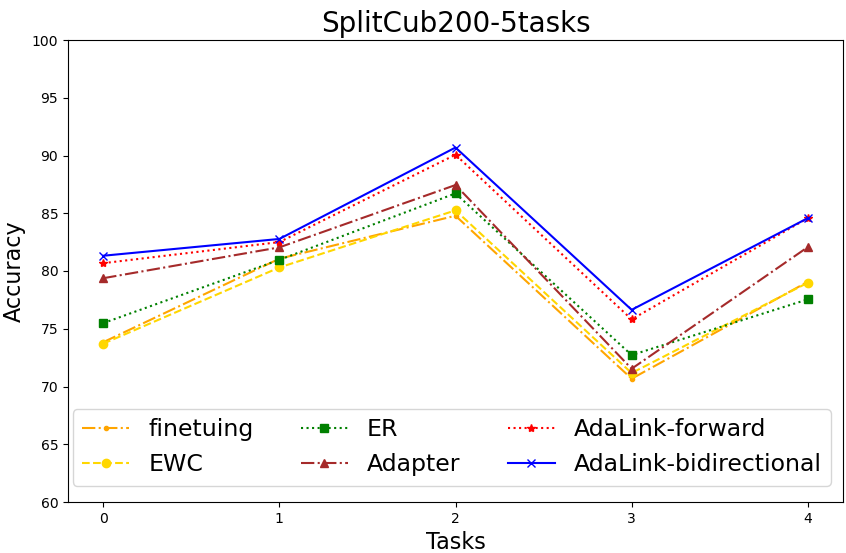}
        \caption{}
        \label{fig:sub2}
    \end{subfigure}
    \caption{ Comparison between baselines and \textit{AdaLink} over average test accuracy of individual tasks. In the above figure, on the X-axis task numbers are mentioned and, on the Y-axis, the average test accuracy of individual tasks is presented.}
    \label{fig:overall}
    \vspace{-0.5cm}
\end{figure*}

The primary aim of \textit{AdaLink-Bidirectional} was to facilitate the transfer of knowledge back to the initial tasks that may not have received sufficient attention during the training phase. In Figure \ref{fig:sub1} and Figure \ref{fig:sub2}, we can observe the improvement in the accuracy of the initial tasks and the ability to effectively transfer knowledge from subsequent tasks back to the initial tasks.

In the experiments detailed in Table \ref{table:1}, approaches that employ adapters have a parameter growth of approximately 2\% for every task compared to the parameters present in the $ViT\_B\_16$. While \textit{MLP} introduced for generating attention weights only adds 0.000034\% additional parameters across all tasks compared to the $ViT\_B_\_16$ which is nearly negligible. We maintained a single MLP across all tasks. The architecture of the MLP consists of three fully connected layers with dimensions 64×32, 32×16, and 16×12 and we choose a task-specific embedding of size 32. Results reported in Table \ref{table:1} for Linked adapters and other baselines are run for 1 epoch. For training the model on Cifar-100 with 10 splits, standalone adapters took around 6 minutes whereas Linked adapters took 7 minutes approximately using a single NVIDIA Tesla V100 GPU.

\section{Ablation Study}

\subsection{Constant Attention Weights}
 To demonstrate the effectiveness of leveraging the MLP in generating attention weights, we experimented on models \textit{AdaLink-Forward-k} and \textit{AdaLink-Bidirectional-k} that use constant attention weights ($\beta=k$), with the value of k=1 determined through validation data.
 
  \setlength{\topsep}{0pt}
\begin{wrapfigure}{r}{0.5\textwidth}
  \centering
  \vspace{-1em}
  \includegraphics[width=\linewidth, height=4cm]{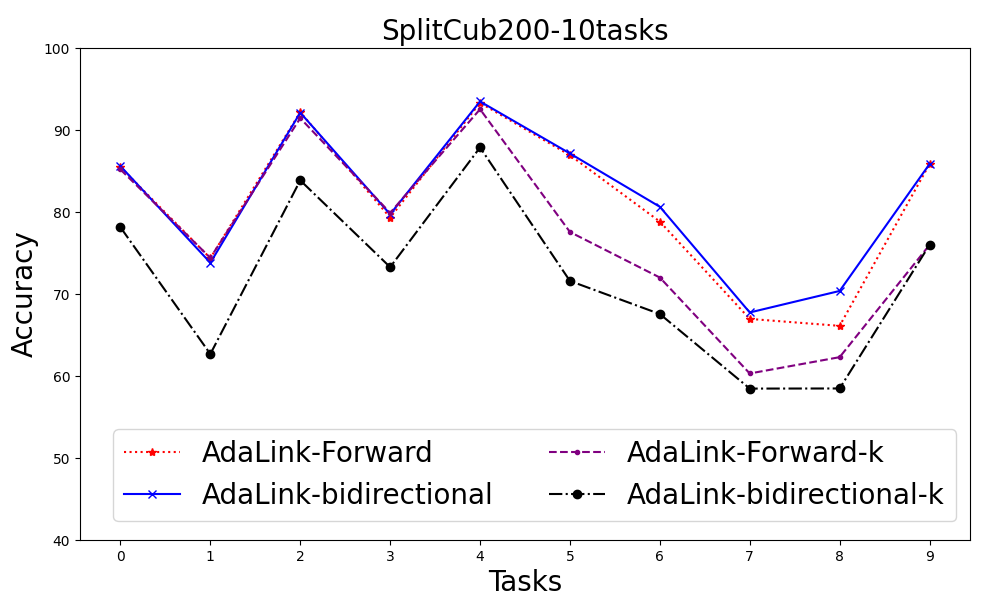}
  \setlength{\intextsep}{0pt}
  \setlength{\abovecaptionskip}{0pt}
  \caption{Comparison between \textit{AdaLink} with MLP and \textit{AdaLink} with constant attention weights (\textit{AdaLink-Forward-k} and \textit{AdaLink-Bidirectional-k}) over average test accuracy of individual tasks on Cub 200 dataset with 10 tasks.}
  \label{fig:constant}
   \vspace{-2em}
\end{wrapfigure}
 Analyzing Fig \ref{fig:constant}, we observe a notable trend, \textit{AdaLink-Forward-k} experiences a significant accuracy drop after the initial tasks, indicating challenges in learning the optimal degree of information flow from other tasks. Similarly, in \textit{AdaLink-Bidirectional-k}, the considerable accuracy drop arises from the use of constant attention weights for all tasks. This strategy overlooks the diverse information flow, as optimal performance requires higher attention weights for similar tasks and lower weights for dissimilar ones. To address this, we employed an MLP that effectively adapts to address the challenge of the degree of information flow across tasks, demonstrating its crucial role in knowledge transfer during both training and testing.

\section{Conclusion}
This paper presents \textit{Linked Adapters}- a novel approach for continual learning that addresses knowledge transfer in PLMs. Our proposed approach, \textit{MLP} based \textit{Linked Adapters} establishes systematic lateral connections between task adapters to enable knowledge transfer. The incorporation of attention weights, guided by a \textit{MLP}, addresses the challenge of determining information flow between tasks. During training, the \textit{MLP} learns to generate attention weights, enabling the current task model to selectively gain knowledge from preceding tasks. During testing, the trained \textit{MLP} proves instrumental in predicting attention weights for subsequent tasks, facilitating knowledge transfer without additional training. Experiments demonstrated the effectiveness of \textit{Linked Adapters} as a flexible and effective approach for continual learning with pre-trained transformer models. Future work will explore conducting experiments in other domains, such as natural language processing (NLP) and vision-language tasks, to assess the broader applicability of the proposed method.

%
%
%
\bibliographystyle{splncs04}
\bibliography{egbib.bib}

\end{document}